# Integrating remote sensing data assimilation, deep learning and large language model for interactive wheat breeding yield prediction

Guofeng Yang[a], Nanfei Jin[b], Wenjie Ai[a], Zhonghua Zheng[c], Yuhong He[d], Yong He[a*]

**Affiliations**
  [a] *College of Biosystems Engineering and Food Science, Zhejiang University, Hangzhou, Zhejiang, China*
  [b] *College of Agriculture and Biotechnology, Zhejiang University, Hangzhou, Zhejiang, China*
  [c] *Department of Earth and Environmental Sciences, The University of Manchester, Manchester, United Kingdom*
  [d] *Department of Geography, Geomatics and Environment, University of Toronto, Mississauga, L5L 1C6, ON, Canada*
  [*]Correspondence: yhe@zju.edu.cn

**ABSTRACT**
Yield is one of the core goals of crop breeding. By predicting the potential yield of different breeding materials, breeders can screen these materials at various growth stages to select the best performing. Based on unmanned aerial vehicle remote sensing technology, high-throughput crop phenotyping data in breeding areas is collected to provide data support for the breeding decisions of breeders. However, the accuracy of current yield predictions still requires improvement, and the usability and user-friendliness of yield forecasting tools remain suboptimal. To address these challenges, this study introduces a hybrid method and tool for crop yield prediction, designed to allow breeders to interactively and accurately predict wheat yield by chatting with a large language model (LLM). First, the newly designed data assimilation algorithm is used to assimilate the leaf area index into the WOFOST model. Then, selected outputs from the assimilation process, along with remote sensing inversion results, are used to drive the time-series temporal fusion transformer model for wheat yield prediction. Finally, based on this hybrid method and leveraging an LLM with retrieval augmented generation technology, we developed an interactive yield prediction Web tool that is user-friendly and supports sustainable data updates. This tool integrates multi-source data to assist breeding decision-making. This study aims to accelerate the identification of high-yield materials in the breeding process, enhance breeding efficiency, and enable more scientific and smart breeding decisions.

# 1. Introduction

According to the *World Population Prospects 2024* report, the global population is projected to increase by 2 billion in the next 60 years, reaching a peak of approximately 10.3 billion by the mid-2080s [1]. The current rate of growth in the yields of major crops remains insufficient to meet the demand driven by this population increase [2]. While conventional breeding methods have significantly improved crop yields and quality [3], [4], the enhancement of yields in existing crops faces substantial challenges, particularly in the context of climate change. This is especially evident in staple crops such as wheat, where the gap between expected yields and future demand continues to widen [5], [6].

Unmanned aerial vehicle (UAV) remote sensing technology, with its high spatial resolution, rapid data acquisition capabilities, and extensive coverage, in combination with multispectral, hyperspectral, and LiDAR technologies, enables precise crop yield prediction [7], [8], [9]. Existing studies on wheat yield prediction can generally be categorized into three methods: process-based model-driven, data-driven, and hybrid. Process-based model-driven methods (e.g., WOFOST [10], AquaCrop [11], APSIM [12]) construct crop growth models that offer good interpretability; however, they rely on simplified assumptions and struggle to address complex and dynamic environmental conditions. Moreover, these models may lack accuracy in predictions when data is scarce. In contrast, data-driven methods (e.g., GRU [13], LSTM [14], Transformer [15]) analyze large-scale datasets to identify complex nonlinear relationships, with prediction accuracy improving as more data becomes available. However, these methods typically suffer from poor interpretability and high data quality requirements, with missing or biased data potentially significantly affecting outcomes. While both methods have their respective strengths and weaknesses, they face limitations in addressing the complexities of agricultural environments. Recently, hybrid methods combining data assimilation with machine learning (deep learning) have gradually emerged as an effective solution to overcome these challenges, due to their exceptional accuracy and adaptability [16], [17], [18]. Data assimilation technology (e.g., EnKF, 4D-Var) can integrate observational data with model predictions to correct model errors, thus improving yield prediction accuracy [19]. When combined with deep learning methods (e.g., CNN, LSTM), they can further enhance the accuracy and stability of yield predictions by uncovering latent nonlinear relationships from large datasets [20]. Hybrid methods offer greater flexibility and robustness when dealing with dynamic agricultural environments, addressing the limitations of individual methods [21]. However, existing hybrid methods still have shortcomings in several areas. For instance, current assimilation algorithms face challenges related to computational efficiency and accuracy instability when handling high-dimensional, nonlinear, and time-varying agricultural data [22]. Moreover, deep learning algorithms often lack effective mechanisms to integrate model outputs with observational data when combined with traditional assimilation methods [23]. Therefore, there is an urgent need to develop new hybrid methods, particularly more efficient assimilation algorithms and deep learning architectures to further enhance the accuracy and adaptability in complex environments.

Existing studies have developed integrated platforms for breeding phenotypic and genetic association, such as Crop-GPA [24], the Smart Breeding Platform [25], and GoldenSeed [26]; breeding data management systems like BreedingEIS [27] and GridScore [28]; and gene analysis and prediction tools like the SoyDNGP website [29] and PidTools [30]. These platforms provide some degree of support for breeding research. However, there remains a lack of user-friendly and convenient tools that utilize remote sensing and phenotyping data directly for yield prediction and analysis [31]. Moreover, breeders face challenges in accurately evaluating the performance of the same or different varieties across various breeding experiments conducted in different regions when comparing breeding materials. They often rely on manually searching for publicly available yield data, a cumbersome and time-consuming process that reduces the efficiency of breeding decisions and may hinder the promotion of potentially superior varieties. Additionally, inconsistencies in field experiment designs and data collection standards make it difficult to directly and effectively compare yield outcomes from different breeding studies [32].

Against this backdrop, the rapid advancement of large language models (LLMs) and retrieval-augmented generation (RAG) technologies has opened new opportunities for intelligent decision support in the field of breeding [33]. In particular, progress in Transformer architectures and self-supervised learning has significantly enhanced the ability of LLMs to process multimodal data and understand complex contexts [34]. LLMs have demonstrated tremendous application potential across various domains, such as content creation, intelligent customer service, and conversational systems, significantly improving user experience [35], [36], [37]. When combined with RAG technology, LLMs can meet real-time data update demands by integrating information retrieval and text generation to deliver more accurate and timely predictions and analyses [38], [39]. This combination of technologies not only enables diverse and personalized interactions but also effectively integrates information from remote sensing data, phenotyping data, and breeding experiment results. It provides breeders with accurate and up-to-date decision support on crop germplasm characteristics, yield performance, cultivation techniques, and suitable planting areas [40].

This study integrates a newly developed hybrid method (remote sensing data assimilation and deep learning) with a large language model to achieve interactive and accurate predictions of wheat breeding yields. Specific goals include: 1. Using the newly designed data assimilation algorithm (WW-4VES) to assimilate the leaf area index (LAI) into the WOFOST model to improve the efficiency and accuracy of data assimilation; 2. Using the partial output results after assimilation and the results of remote sensing inversion to drive the time series temporal fusion transformer (TFT) model to predict wheat yield, and explore its yield prediction performance in different regions and natural weed competition environments; 3. Based on the above yield prediction method and using LLM and RAG technology, build a friendly and continuously updated yield prediction interactive Web tool to integrate multi-source data to support breeding decisions. Through the realization of these goals, the identification of high-yield materials in the breeding process can be accelerated, breeding efficiency can be improved, and scientific and smart breeding decisions can be realized.

## 2. Data

*2.1. Study area and experiment design*

The study area was set up in Huzhou and Hangzhou city, Zhejiang Province, China (Changxing Agricultural Experimental Station of Zhejiang University in Huzhou, Yuhang Agricultural Experimental Station of Zhejiang University in Hangzhou, and Zijingang Agricultural Science and Technology Innovation Experimental Center of Zhejiang University in Hangzhou) (Fig. 1, Table S1). The experiment lasts for three years (2021-2024), with 305 wheat germplasms, 351 wheat germplasms, and 60 high-generation wheat lines. The wheat germplasms come from many countries around the world, and the high-generation wheat lines come from the wheat breeding group of Zhejiang University (Table S2). Specific field cultivation, fertilization, and irrigation information are shown in Table S3. No pesticides and herbicides were sprayed during the wheat growth period, and no manual or mechanical weeding was performed. The natural growing environment was maintained in this study in order to breed wheat varieties that have greater competitive advantages over weeds. The study areas are characterized by a subtropical monsoon climate that is warm and humid, with hot, rainy summers and cold, dry winters.

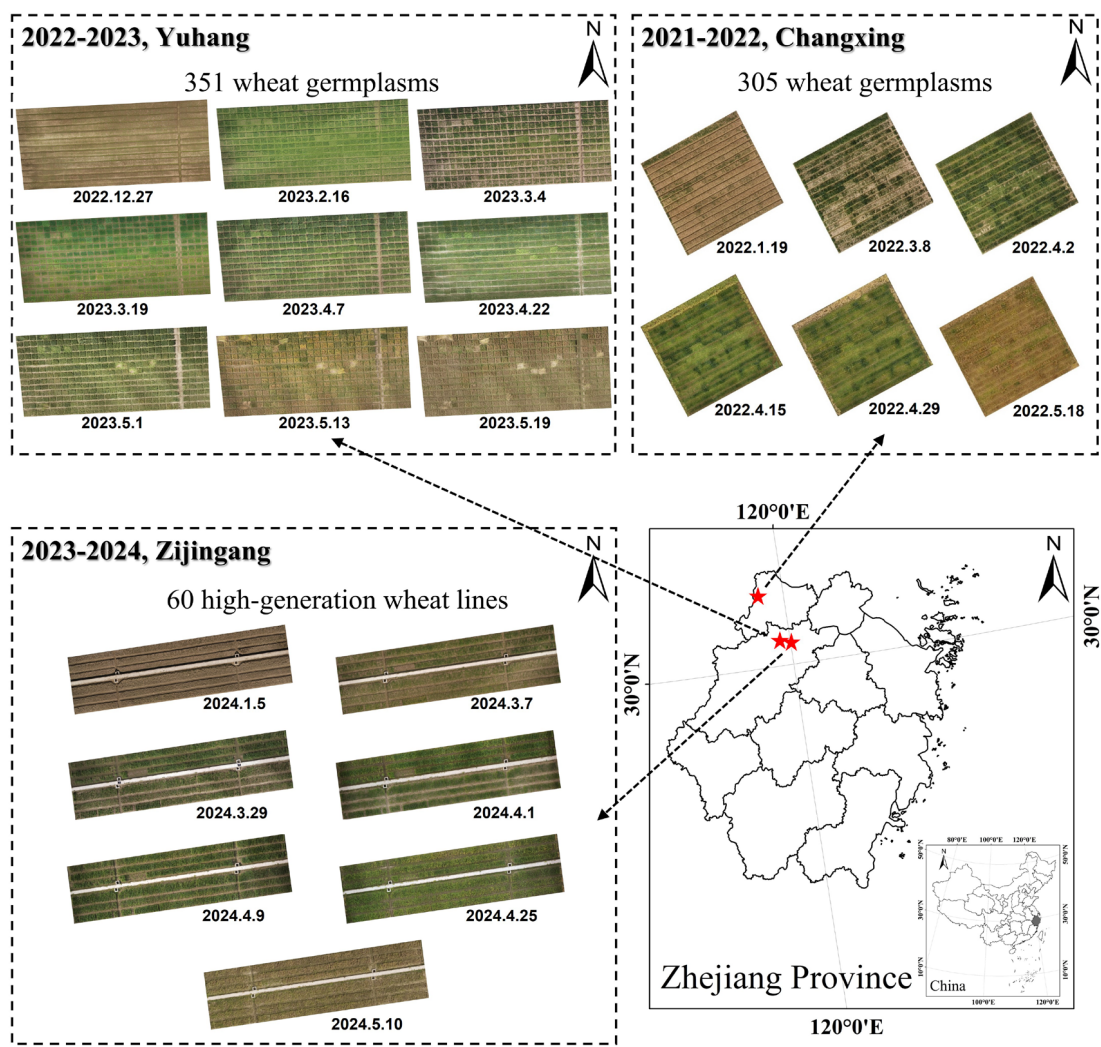

**Fig. 1.** Location of study areas and time series changes of wheat growth in the field.

*2.2. Data collection and preprocessing*

*2.2.1. Remote sensing, phenotypic and environmental data*

This study used the Matrice 300 RTK (DJI, China) equipped with the Zenmuse L1 (DJI, China) LiDAR and the Phantom 4 Multispectral (DJI, China) to acquire three years of point cloud and multispectral data. The data were collected on the same day, under clear, cloudless skies, and low wind conditions, at a flight altitude of 15m (Fig. 2a). The UAV remote sensing data collection included six times from 2021 to 2022, nine times from 2022 to 2023, and seven times from 2023 to 2024 (Fig. 2b). For phenotyping data, LAI and canopy height (CH) of wheat were collected only five times in 2021-2022, and the wheat grain yield was not obtained; LAI and CH of wheat were collected a total of nine times in 2022-2023, and the wheat grain yield was obtained; LAI and CH of wheat were not collected in 2023-2024, and only UAV remote sensing data were collected seven times, and the wheat grain yield was obtained (Fig. 2b). LAI was measured using the LAI-2000C analyzer, with five evenly distributed sampling points per plot, and the measurements were repeated three times to obtain an average. CH was manually measured using a folding ruler to determine the height of five wheat plants, and the average height was taken. Yield refers to the grain yield of wheat after threshing, expressed in kg/ha. The meteorological data for the three growing seasons included daily average temperature, dew point temperature, precipitation, net solar radiation intensity, and wind speed. These data were sourced from meteorological instruments at the agricultural experimental station and the China Meteorological Administration. The soil data for the three-year study area included pH, total nitrogen, available potassium, available phosphorus, and soil organic carbon. Soil samples (0-20 cm) were collected and sent to the soil testing company for analysis (Fig. 2c).

*2.2.2. Yield database*

The wheat yield database has 6,253 entries, which are from the China Seed Industry Big Data Platform (202.127.42.145/bigdataNew/home/index). These data are obtained through the variety of approval trials conducted by the National Crop Variety Approval Committee of China, following the relevant provisions of the *Seed Law* and the *Main Crop Variety Approval Measures*. It is representative and authoritative, and the data is dynamically updated (Fig. 2d). Each entry in the database includes the approval number, variety name, applicant, breeder, variety source, characteristics, yield performance, key points of cultivation technique, and suitable planting area (Table S4).

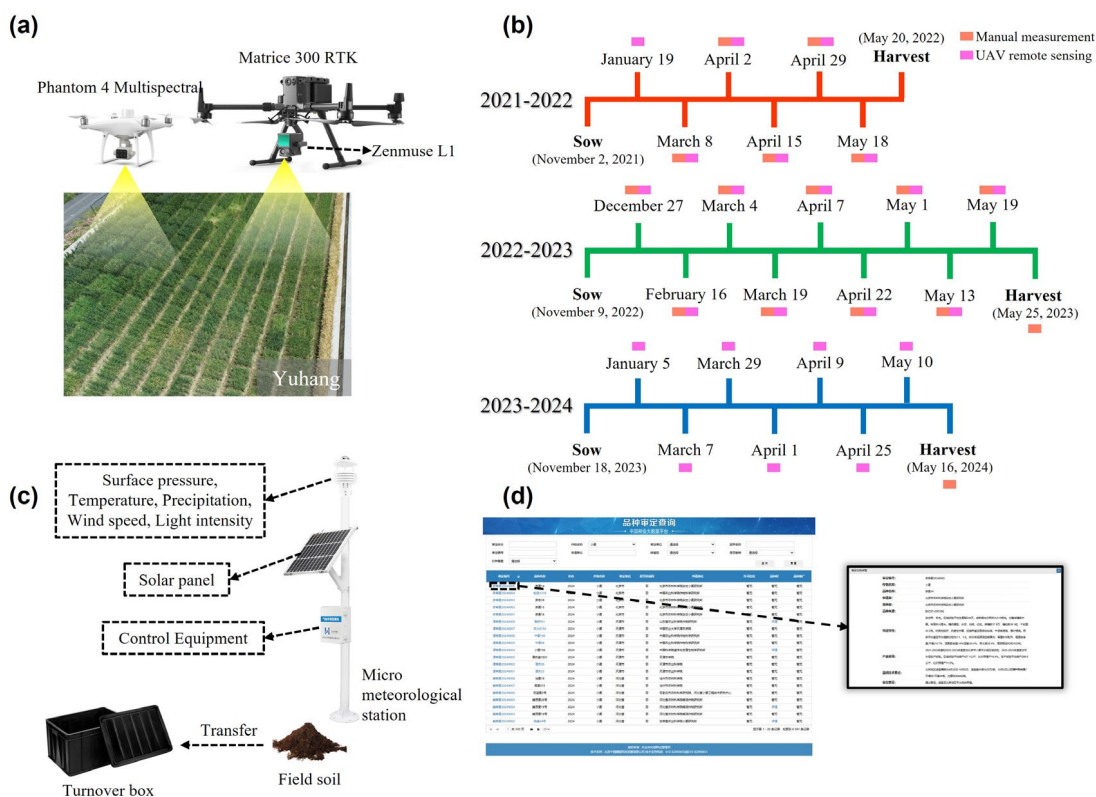

**Fig. 2.** Data collection. (a) Using UAVs to obtain wheat remote sensing data. (b) Time for UAV remote sensing and manual measurement. (c) Collection of meteorological and soil data. (d) Wheat variety approval query results on the China Seed Industry Big Data Platform, which include yield performance and other data of different wheat varieties.

*2.2.3. Preprocessing*

This study utilized Pix4Dmapper software (Pix4D, Switzerland) to process multispectral data. Radiometric calibration was performed using images of a radiometric calibration board to correct spectral data across different bands, ensuring data accuracy and consistency. The calibrated data were used to generate digital orthophoto maps (DOMs), which were subsequently normalized to create reflectance images. These reflectance images were employed to calculate both conventional and novel vegetation indexes (VIs) for yield prediction (Table 1). Additionally, Terra software (DJI, China) was employed to create a bare-earth digital elevation model (DEM) based on LiDAR point cloud. Using point cloud acquired at multiple times, digital surface models (DSMs) were constructed. A Python script was developed to perform pixel-by-pixel subtraction of the DEM from the DSM, enabling the calculation of CH in wheat-growing areas. To validate the feasibility of the CH calculation method, remotely sensed CH values were compared with field-measured CH values. The results demonstrated strong correlations (Changxing: $R^2$ = 0.863, RMSE = 0.105 m; Yuhang: $R^2$ = 0.837, RMSE = 0.136 m), confirming the accuracy and reliability of the method. Furthermore, the canopy volume (CV) of individual plots was measured using the measurement tool of Terra software. Specifically, CV was calculated based on point cloud using both the lowest point plane and the mean plane as reference baselines. The average of the two values was taken as the final CV.

To support breeding research using only remote sensing data without manual collection of wheat LAI measurements, this study exclusively collected remote sensing data in the Zijingang experiment. First, inversion models were constructed using manually collected LAI data and corresponding VIs derived from UAV-based remote sensing in the Changxing and Yuhang experiments (Table 1). Several methods, including support vector regression (SVR), random forest regression (RFR), XGBoost, and LightGBM, were developed for this purpose. The dataset was randomly split into training set, validation set and test set in a ratio of 3:1:1. The performance of inversion of different methods is compared on the test set. Results show that XGBoost outperformed other methods in both experiments (Table S5). Referring to related research [41], the LAI inversion accuracy achieved by the XGBoost method satisfies the precision requirements for assimilation with the crop growth model WOFOST and remote sensing data. Finally, a new XGBoost model was trained using the full set of LAI and VI data from both experiments. This model was applied to the Zijingang experiment, where LAI data were not manually collected, to invert LAI for different wheat breeding materials.

All data in this study are publicly available (https://doi.org/10.5281/zenodo.14376799).

**Table 1**

Vegetation indexes used in this study.

| Vegetation index | Formulation | Reference |
| --- | --- | --- |
| Chlorophyll Index red-edge (CIre) | $CIre = \frac{NIR}{RE} - 1$ | [42] |
| Chlorophyll Index green (CIg) | $CIg = \frac{NIR}{G} - 1$ | [43] |
| Enhanced Vegetation Index (EVI) | $EVI = \frac{2.5 \times (NIR - R)}{NIR + 6 \times R - 7.5 \times B + 1}$ | [44] |
| Green Normalized Difference Vegetation Index (GNDVI) | $GNDVI = \frac{NIR - G}{NIR + G}$ | [45] |

| Green Re-normalized Different Vegetation Index (GRDVI) | $GRDVI = \frac{NIR - G}{\sqrt{NIR + G}}$ | [46] |
|---|---|---|
| kernel Normalized Difference Vegetation Index (kNDVI) | $kNDVI = tanh\left(\left(\frac{NIR - R}{2\sigma}\right)^2\right)$, where $\sigma$ is a tunable length-scale parameter intended to capture nonlinear sensitivity of NDVI to vegetation density. If $\sigma = 0.5(NIR + R)$, which simplifies to $kNDVI = tanh((NDVI)^2)$. | [47] |
| Plant Senescence Reflectance Index (PSRI) | $PSRI = \frac{R - B}{NIR}$ | [48] |
| Modified Soil Adjusted Vegetation Index (MSAVI) | $MSAVI = \frac{2 \times NIR + 1 - \sqrt{(2 \times NIR + 1)^2 - 8 \times (NIR - R)}}{2}$ | [49] |
| Modified Transformed Chlorophyll Absorption Ratio Index (MTCARI) | $MTCARI = 3 \times ((NIR - RE) - 0.2 \times (NIR - RE)(\frac{NIR}{G}))$ | [50] |
| Modified Triangular Vegetation Index (MTVI) | $MTVI = 1.2 \times (1.2 \times (NIR - G) - 2.5 \times (R - G))$ | [51] |
| Nonlinear Vegetation Index (NLI) | $NLI = \frac{NIR^2 - R}{NIR^2 + R}$ | [52] |
| Normalized Difference Red-edge (NDRE) | $NDRE = \frac{NIR - RE}{NIR + RE}$ | [53] |
| Normalized Difference Vegetation Index (NDVI) | $NDVI = \frac{NIR - R}{NIR + R}$ | [54] |
| Near-Infrared Reflectance of Vegetation (NIRv) | $NIR_v = NIR \times NDVI$ | [55] |
| Normalized NIR (NNIR) | $NNIR = \frac{NIR}{NIR + R + G}$ | [56] |
| Optimizing Soil Adjustment Vegetation Index (OSAVI) | $OSAVI = \frac{1.16 \times (NIR - R)}{NIR + R + 0.16}$ | [57] |
| Ratio Vegetation Index (RVI) | $RVI = \frac{R}{NIR}$ | [58] |
| Soil-Adjusted Vegetation Index (SAVI) | $SAVI = \frac{1.5 \times (NIR - R)}{NIR + R + 0.5}$ | [59] |
| Simple Ratio (SR) | $SR = \frac{NIR}{R}$ | [60] |
| Red-Edge Simple Ratio (SRre) | $SRre = \frac{NIR}{RE}$ | [61] |
| Structure Intensive Pigment Index (SIPI) | $SIPI = \frac{NIR - B}{NIR + R}$ | [62] |
| Triangular Vegetation Index (TVI) | $TVI = 60 \times (NIR - G) - 100 \times (R - G)$ | [63] |
| Visible-band Difference Vegetation Index (VDVI) | $VDVI = \frac{2G - R - B}{2G + R + B}$ | [64] |
| Visible Atmospherically Resistant Index (VARI) | $VARI = \frac{G - R}{G + R - B}$ | [65] |
| Wide Dynamic Range Vegetation Index (WDRVI) | $WDRVI = \frac{0.1 \times NIR - R}{0.1 \times NIR + R}$ | [66] |

## 3. Method

### 3.1. Hybrid method

The hybrid method in this study integrates the WOFOST model, the WW-4VES data assimilation algorithm, and the TFT model. The inputs of the hybrid method are different data collection dates and corresponding development days (DVDs), remote sensing inversion data (VIs, CV, CH) and LAI values. Before using the hybrid method, the crop model must be calibrated and validated (Section 3.1.1). Subsequently, the newly designed WW-4VES data assimilation algorithm is employed to assimilate LAI into the WOFOST model (Section 3.1.2). Finally, the partially assimilated outputs and the remote sensing inversion results are used to drive the time series TFT model to predict wheat yield (Section 3.1.3).

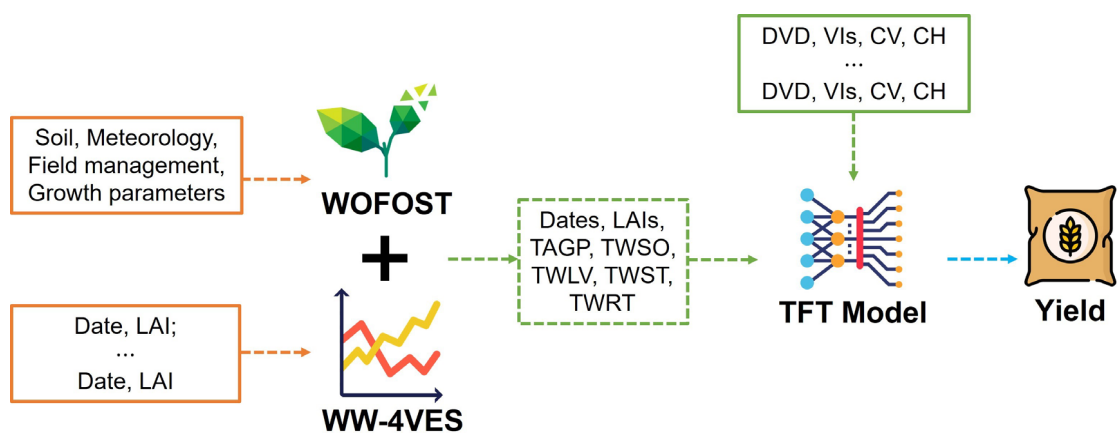

**Fig. 3.** Hybrid method in this study.

*3.1.1. WOFOST model calibration and validation*

The WOFOST (WOrld FOod STudies) model is a dynamic process model used to simulate crop growth. The WOFOST model estimates crop biomass accumulation, yield, and water requirements by simulating various physiological processes in crop growth, including photosynthesis, respiration, and transpiration. This study refers to the latest document "A gentle introduction to WOFOST" on the WOFOST official website and determines the range of parameters that need to be calibrated in WOFOST (Table S6). "Free" parameters are calibrated according to measured data; "Additional" parameters can be calibrated as needed, using the WOFOST model default values or related reference values; "Static" parameters that are unlikely to require calibration use the WOFOST model default values. The Python Crop Simulation Environment (PCSE) framework provides an environment for running the WOFOST crop growth model.

According to the WOFOST parameter calibration scheme (Table S6), the parameters of the three experiments in this study are calibrated separately. After parameter calibration, the simulated LAI or yield of each experiment had a good fit with the measured LAI or yield overall (Table S7), indicating the effectiveness of the WOFOST model in predicting wheat growth. Furthermore, although the predicted LAI or yield fit well with the measured LAI or yield, there are large regional differences in the values of LAI or yield, indicating that it is necessary to use remote sensing observation data to assimilate the results of crop model prediction.

*3.1.2. WW-4VES data assimilation algorithm*

The LAI of crops is one of the important parameters related to the growth status and development process of crops. Therefore, the LAI of crops is selected as the update state to achieve data assimilation and yield prediction. First, the driving parameters are input into the WOFOST model, including daily meteorological data, crop parameters, soil parameters, and agricultural management practice parameters. The WOFOST model is then executed to generate the predicted LAI at the corresponding time, serving as the background field LAI. Subsequently, combined with the observed field LAI (LAI measured by ground instruments or LAI obtained by remote sensing inversion during wheat growth), the WW-4VES data assimilation algorithm fused the observed LAI with the LAI predicted by the model. Since both field measurement data and remote sensing data may contain errors, quality control and estimation of observational errors are necessary. Assuming that observational errors follow a Gaussian distribution, the mean and variance of the obtained LAI are applied to the Gaussian distribution formula to optimize model uncertainty and enhance predictive accuracy. Throughout the entire growth season of wheat, the above assimilation process is repeated, with each observed LAI value assimilated into the WOFOST model. This sequentially corrects the state of the model, generating a simulated growth process that increasingly approximates the actual growth. At the end of the growth season, the WOFOST model outputs the final yield, which is represented as the TWSO. In addition, this study applies hyperparameter optimization to the WOFOST model using Optuna [67]. Specifically, the Tree-structured Parzen Estimator (TPE) sampler and the Hyperband pruner are used iteratively to minimize the cost function for the WOFOST model. The TPE sampler efficiently searches the hyperparameter space to identify the optimal combination of hyperparameters for improving model performance. Hyperband accelerates the hyperparameter optimization process by dynamically adjusting resource allocation. When the cost function reaches its minimum within the specified range, the WOFOST output based on the optimal parameter set is considered the optimal output. In this study, the WW-4VES data assimilation algorithm is implemented in Python. This study also developed the EnKF method [68] and the EnKF-4DVar method [69] to compare the performance of yield prediction, and tried to keep the variables and parameters consistent.

The WW-4VES data assimilation algorithm integrates adaptive weight, variable time windows, four-dimensional variational, and ensemble square root filtering. The derivation process can be summarized in the following steps: 1. State evolution and initial estimation. The initial state is estimated using the background error covariance $P_0$ and observation data. 2. Observation update. The observational increment $\delta y_i$ is computed, and the state is updated using the Kalman gain matrix $K_i$. 3. Adaptive weight adjustment. The adaptive weight $\alpha_i$ is computed based on the background and observation errors to refine the analysis field. 4. Time window adjustment. The time window length $T_w(t+1)$ is dynamically adjusted according to the error growth rate $r(t)$ to accommodate different observational environments.

In the objective function $\mathcal{F}(x_0)$, the first term represents the error of the background field, while the second term represents the error between observations and model states; these terms are combined using a weighted approach based on the error covariance matrices. The objective function $\mathcal{F}(x_0)$ is used to evaluate the appropriateness of the current initial state $x_0$. Let $x_0$ denote the initial state of the current system (subject to optimization), and $x_b$ denote the background field (the prior estimate of the initial state). The matrix $B$ denotes the error covariance of the background field, providing a measure of the magnitude and uncertainty of background errors. $x_i$ denotes the state of the system at the $i$-th time step, and $y_i$ denotes the observation data at this time step. The operator $H_i$ is the observation operator, mapping the model state to the observation space. Finally, $R_i$ denotes the observation error covariance matrix at the $i$-th time step, which quantifies the magnitude of observational errors.

$$\mathcal{F}(x_0) = \frac{1}{2}(x_0 - x_b)^T B^{-1}(x_0 - x_b) + \frac{1}{2}\sum_{i=0}^{N}(y_i - H_i x_i)^T R_i^{-1}(y_i - H_i x_i) \qquad (1)$$

The system state evolves through the model, incorporating an error term $\eta_i$, to produce the background state $x_{i+1}$ at the next time step. Here, $x_{i+1}$ denotes the system state at the $i+1$-th time step. The state evolution operator $M_i$ describes how the state at time $i$-th is used to calculate the state at time $i+1$-th. The error term $\eta_i$ denotes model error, capturing the uncertainty and inaccuracies in the state evolution process.

$$x_{i+1} = M_i(x_i) + \eta_i \qquad (2)$$

The ensemble mean $\bar{x}_0$ denotes the average initial state across multiple sample states, while the covariance matrix $P_0$ describes the error and uncertainty associated with these initial states. Specifically, $\bar{x}_0$ is the mean of the ensemble samples, providing an average of the initial states. $K$ denotes the number of ensemble samples, and $x_0^{(k)}$ denotes the initial state of the $k$-th ensemble sample. The matrix $P_0$ is the error covariance matrix of the background field, reflecting the uncertainty of the background field.

$$\bar{x}_0 = \frac{1}{K}\sum_{k=1}^{K} x_0^{(k)} \qquad (3)$$

$$P_0 = \frac{1}{K-1}\sum_{k=1}^{K}(x_0^{(k)} - \bar{x}_0)(x_0^{(k)} - \bar{x}_0)^T \qquad (4)$$

The observation increment $\delta y_i$ reflects the discrepancy between observational data and the model state, serving as a key variable in the subsequent assimilation update. Specifically, $\delta y_i$ denotes the difference between observational data and the model state in the observation space. Here, $y_i$ denotes the observational data at the $i$-th time step, $H_i$ is the observation operator that projects the model state $x_i$ into the observation space, and $x_i$ is the model state at the $i$-th time step.

$$\delta y_i = y_i - H_i x_i \qquad (5)$$

The state update formula uses the Kalman gain $K_i$ to adjust the background field $x_{b,i}$ based on the observation increment $\delta y_i$, yielding the updated analysis field $x_{a,i}$. Here, $x_{a,i}$ denotes the analysis state after assimilation (the updated state at the $i$-th time step), while $x_{b,i}$ denotes the background field at the $i$-th time step. The Kalman gain matrix $K_i$ determines the weighted contribution of the background field and observational data.

$$x_{a,i} = x_{b,i} + K_i \delta y_i \qquad (6)$$

The Kalman gain matrix $K_i$ determines the weighting between the background field and the observational data; typically, when the observation error $R_i$ is small, the observational data are given greater weight. Specifically, $K_i$ is the Kalman gain matrix at the $i$-th time step, $P_i$ is the background error covariance matrix at the $i$-th time step, $H_i$ is the observation operator that projects the model state into the observation space, and $R_i$ is the observation error covariance matrix, which quantifies the magnitude of observation errors.

$$K_i = P_i H_i^T (H_i P_i H_i^T + R_i)^{-1} \qquad (7)$$

The adaptive weight $\alpha_i$ is dynamically adjusted based on the magnitudes of the observation and background errors to better balance their respective influences. Here, $\alpha_i$ is the adaptive weight used to balance the uncertainties in the observations and the background field. The $\text{tr}(R_i)$ is the trace of the observation error covariance matrix $R_i$, denoting the overall magnitude of observation errors, and $\text{tr}(P_i)$ is the trace of the background error covariance matrix $P_i$, denoting the overall magnitude of background errors.

$$\alpha_i = \frac{1}{\text{tr}(R_i) + \text{tr}(P_i)} \qquad (8)$$

With the inclusion of the adaptive weight $\alpha_i$, the final analysis state $x_{a,i}$ is determined by both the background field $x_{b,i}$ and the observation increment $\delta y_i$, adjusted by the Kalman gain $K_i$. Here, $x_{a,i}$ is the analysis state after assimilation, $\alpha_i$ is the adaptive weight that adjusts the weighting between the background field and observational data, and $K_i$ is the Kalman gain matrix.

$$x_{a,i} = (1 - \alpha_i)x_{b,i} + \alpha_i K_i \delta y_i \qquad (9)$$

The dynamic adjustment of the time window is controlled by the error growth rate $r(t)$; the window is shortened when background errors are large and extended when observation errors are large. Here, $T_w(t+1)$ denotes the time window length at the next time step, and $T_w(t)$ denotes the current time window length. $r(t)$ is the error growth rate at the current time step, while $r_0$ is the preset threshold for the error growth rate, used to determine if an adjustment to the time window length is needed. The parameter $\beta$ is an adjustment coefficient that controls the sensitivity of window changes.

$$r(t) = \frac{\text{tr}(P_i)}{\text{tr}(R_i)} \qquad (10)$$

$$T_w(t+1) = T_w(t) \times \left(\frac{r(t)}{r_0}\right)^\beta \qquad (11)$$

The above formulas (1-11) are computed in sequence through multiple steps to finally determine the assimilation state of the system and dynamically adjust the time window length to adapt to different observation and model environments. Finally, we input the partial outputs (TAGP, TWSO, TWLV, TWST, TWRT) of the combined WOFOST and WW-4VES data assimilation algorithm, different data collection dates and corresponding LAI values into the subsequent TFT model.

*3.1.3. Time Series TFT Model*

This study uses the TFT model [70], which is a deep learning model specifically for time series forecasting. It is designed to process multivariate time series data and provide efficient and interpretable forecasts. The data used in the TFT model include the partial outputs (TAGP, TWSO, TWLV, TWST, TWRT) after assimilated, different data collection dates and corresponding DVDs, remote sensing inversion data (VIs, CV, CH) and LAI values. The dataset was randomly split into training set, validation set and test set in a ratio of 3:1:1. This study implemented the TFT model using Python and PyTorch Forecasting package. Optuna was used to optimize the hyperparameters of the TFT model [67]. The output of the TFT model under the optimal parameters was the final predicted yield of this study. To compare the effect of yield prediction, this study used Keras (https://keras.io/) and Scikit-learn [71] to develop Long Short-Term Memory (LSTM), Gated Recurrent Unit (GRU) and Transformer networks, and tried to keep the network layers, parameters, training and prediction consistent.

*3.2. Interactive yield predict tool*

LlamaIndex (https://www.llamaindex.ai/) uses efficient information retrieval and indexing techniques to integrate external data with LLM, supporting RAG to enhance the accuracy and richness of generated results. Therefore, this study developed an application combining RAG technology and LLM based on the LlamaIndex framework, i.e., an interactive yield prediction Web tool (ChatYield) (Fig. 4c). The LLM selects the open source Llama3.2-3.2-3B-Instruct (https://www.llama.com/), and the vector database selects Milvus (https://milvus.io/). Specifically, the data in the yield database are converted, and embedding vectors are generated using BGE-M3 [72], which are then stored in the Milvus vector library. Then, LlamaIndex receives user input and initiates a search request to the Milvus database based on the input, through the pipeline of hybrid search (BGE-M3) and reranking (BGE-Reranker-V2-M3). LlamaIndex merges the retrieval results with the input to form a new prompt. Afterwards, LlamaIndex inputs the new prompt into Llama3.2-3.2-3B-Instruct, and Llama3.2-3.2-3B-Instruct implements answer generation based on the new prompt.

Unlike the typical interaction processes between users and LLMs (Fig. 4a) or users and systems combining LLMs with RAG technologies (Fig. 4b), the LLM of ChatYield simultaneously processes outputs from the hybrid method and results retrieved from the yield database via RAG, generating final yield prediction responses (Fig. 4c). When predicting the yield of breeding material, the user inputs the variety name, specific data collection date, DVD, remote sensing inversion data (VIs, CV, CH), and LAI value. The hybrid method then generates a yield prediction based on these inputs, while ChatYield retrieves matching yield-related data from the database. Finally, the LLM of ChatYield integrates the output of the hybrid method with RAG-retrieved data to produce the final yield prediction response.

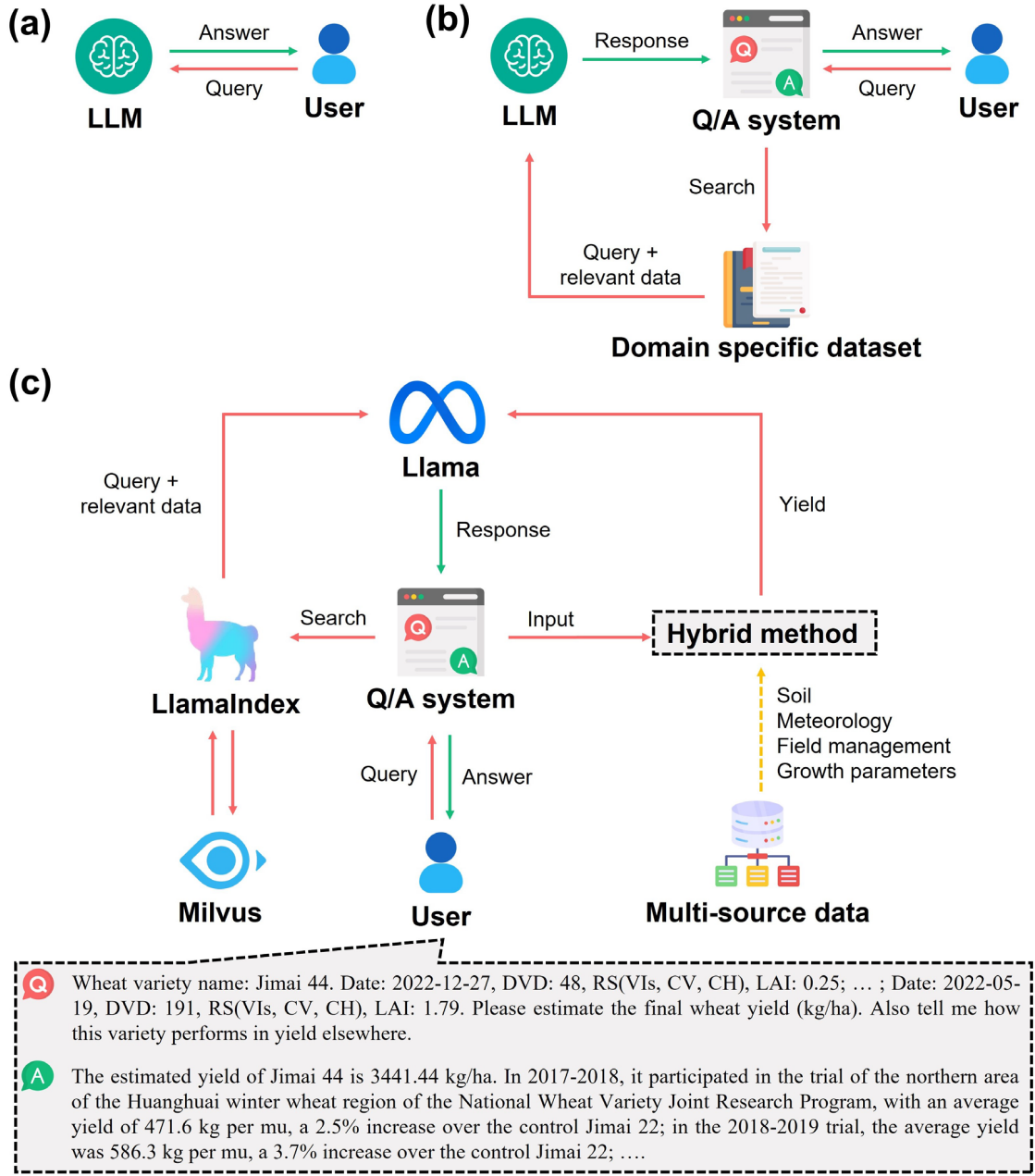

**Fig. 4.** Interaction process. (a) Typical interaction process between user and LLM. (b) Typical interaction process between user and system combining LLM with RAG technology. (c) Interaction process of ChatYield.

*3.3. Evaluation of the accuracy of yield predict*

The coefficient of determination ($R^2$) and root mean square error (RMSE) are used as evaluation indicators to evaluate the effect of yield prediction.

$$R^2 = \frac{\sum_{i=1}^{n}(x_i - \bar{x})^2 (y_i - \bar{y})^2}{\sum_{i=1}^{n}(x_i - \bar{x})^2 \sum_{i=1}^{n}(y_i - \bar{y})^2} \tag{12}$$

In the formula, $x_i$, $y_i$, $\bar{x}$, $\bar{y}$ are the measured wheat yield, predicted yield, average measured yield and average predicted yield, respectively, and $n$ is the number of samples. Generally, the closer $R^2$ is to 1, the better the fit between the predicted wheat yield and the measured yield.

$$\text{RMSE} = \sqrt{\frac{1}{n}\sum_{i=1}^{n}(y_i - x_i)^2} \qquad (13)$$

Among them, $x_i$ and $y_i$ are the measured and predicted wheat yields, respectively, and $n$ is the number of samples.

## 4. Results

*4.1. Impact of Data Assimilation on Predicted Yields*

To assess the specific impact of data assimilation, the wheat yield predictions before and after data assimilation were compared (Fig. 5). We found that the predictions of the crop model without assimilation were significantly different from the measured yield. In contrast, the predicted yield of the crop model after assimilation was more consistent with the measured yield. For the Yuhang experiment, the $R^2$ and RMSE before assimilation were 0.652 and 538.9 kg/ha, respectively (Fig. 5a). After assimilation using the WW-4VES data assimilation algorithm, the yield prediction was further improved, with the $R^2$ increase of 0.164 and the RMSE decrease of 147.4 kg/ha (Fig. 5b). For the Zijingang experiment, the $R^2$ and RMSE before assimilation were 0.607 and 690.6 kg/ha, respectively (Fig. 5c). After assimilation, the yield prediction was further improved, with the $R^2$ increase of 0.075 and the RMSE decrease of 68.8 kg/ha (Fig. 5d). These results demonstrate that our WW-4VES data assimilation algorithm improves yield predictions by assimilating LAI dynamics into the WOFOST crop model.

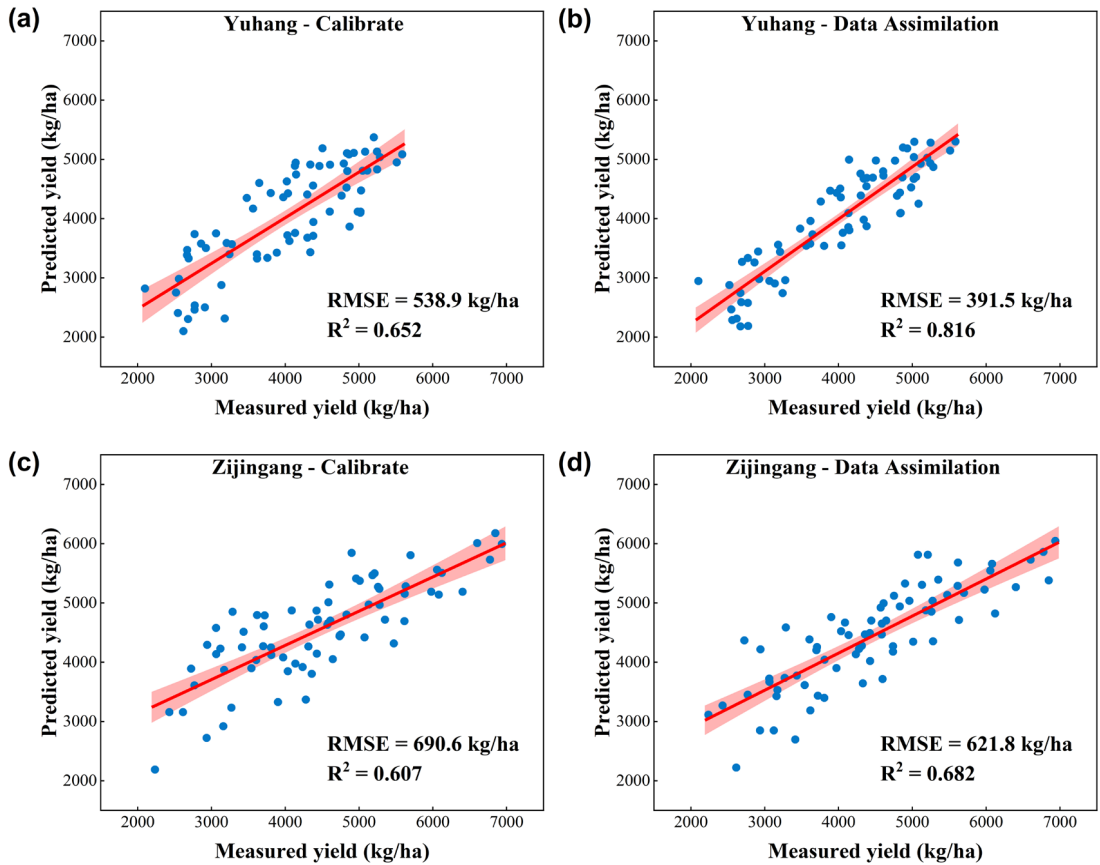

**Fig. 5.** Measured and predicted yields for the Yuhang and Zijingang experiments. (a) and (c) show

the measured and predicted yields after model calibration but before data assimilation; (c) and (d) show the measured and predicted yields after data assimilation.

*4.2. Yield prediction performance of hybrid method*

This study compares process-based model-driven methods (calibrated WOFOST, calibrated WOFOST with EnKF data assimilation, calibrated WOFOST with EnKF-4DVar data assimilation, calibrated WOFOST with WW-4VES data assimilation), data-driven methods (LSTM, GRU, and Transformer), and our hybrid method. The results show that our hybrid method outperforms both process-based model-driven and data-driven methods in terms of the highest $R^2$ (Yuhang: 0.831; Zijingang: 0.704) and the lowest RMSE (Yuhang: 372.8 kg/ha; Zijingang: 605.3 kg/ha) (Fig. 6). However, the $R^2$ of process-based model-driven and data-driven methods are relatively lower (Yuhang: 0.652 ~ 0.816; Zijingang: 0.607 ~ 0.682), with higher RMSE (Yuhang: 391.5 ~ 538.9 kg/ha; Zijingang: 621.8 ~ 690.6 kg/ha), indicating that these methods have poor yield prediction performance. Therefore, the hybrid method is more suitable for wheat yield prediction.

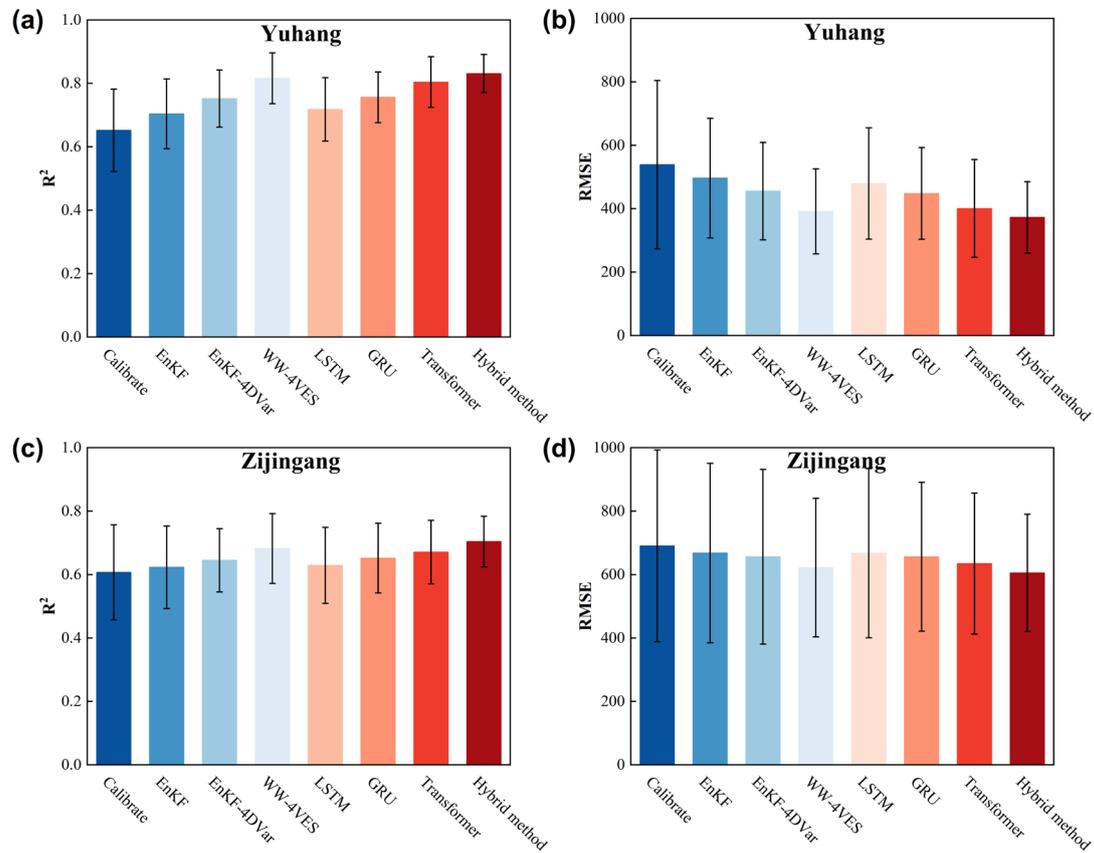

**Fig. 6.** Results of wheat yield prediction using different methods. The error bars indicate the range of maximum and minimum values. (a) and (b) represent the Yuhang experiment, and (c) and (d) represent the Zijingang experiment.

The hybrid method was used to process data containing different numbers of times or orders and to analyze the differences between predicted and measured yields (Fig. 7). When data with fewer number of times are used, the difference between the predicted and measured yields is greater, and the difference gradually decreases as the number of growth stages increases. For each experiment, when all data were used (nine times data were collected for the Yuhang experiment and seven times data were collected for the Zijingang experiment), the predicted yield and the measured yield showed a strong linear fit, with an $R^2$ of 0.831 and an RMSE of 372.8 kg/ha for the Yuhang experiment and an $R^2$ of 0.704 and an RMSE of 605.3 kg/ha for the Zijingang experiment. The results show that more data can usually allow the model to better capture the relationship between features, reduce the bias of the model, and improve the accuracy of predictions. Especially when the training data is noisy or unbalanced, more data can help the model predict yield more accurately. When the total number of times the data is used is fixed, yield prediction accuracy improves when the data includes more from the middle and later stages of growth than data containing more from the early and middle stages. Specifically, when using data collected five times, the prediction results using the last five collected data increased $R^2$ by 0.175 and RMSE by 156.6 kg/ha for the Yuhang experiment compared with the prediction results using the first five collected data. The prediction results using the last three collected data increased $R^2$ by 0.147 and RMSE by 128.7 kg/ha for the Zijingang experiment compared with the prediction results using the first three collected data.

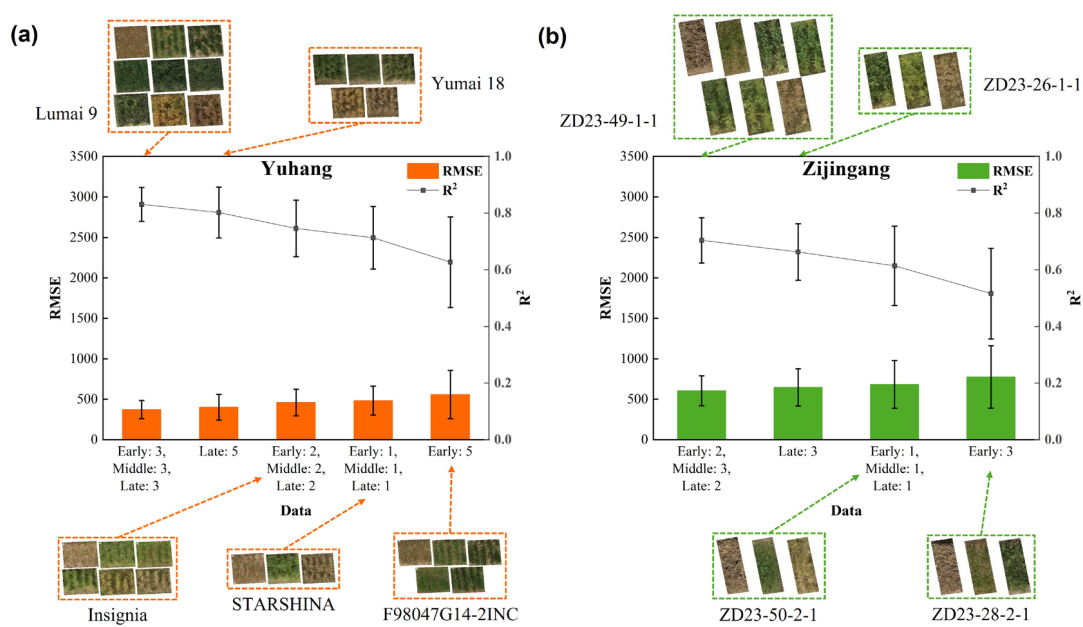

**Fig. 7.** The hybrid method was used to process data containing different numbers of times or orders and to analyze the differences between predicted and measured yields. (a) Results for the Yuhang experiment. (b) Results for the Zijingang experiment. Error bars indicate the range of maximum and minimum values. Each dashed box represents the growth changes of certain wheat germplasm or high-generation wheat lines in this dataset.

The field conditions at maturity are shown in Fig. 8a-c. We use the hybrid method and visualize the spatial distribution of wheat yield predictions for different varieties (Fig. 8g-i). For most breeding materials, the predicted yield was close to the measured yield (Fig. 8e, h; Fig. 8f, i). We compare the yield prediction results using only the data assimilation algorithm WW-4VES (Fig. 8d) and the yield prediction results of the hybrid method (Fig. 8g), and find that the overall yield prediction results are similar, with large differences in the predictions of a few wheat breeding materials. In addition, we found that the average wheat yield of different germplasms from Australia, Turkey and the United States in the Changxing experiment exceeded the local average yield per hectare (Fig. 8j), and the average wheat yield of different germplasms from Japan, Italy, Turkey, Argentina and Australia in the Changxing experiment exceeded the local average yield per hectare

(Fig. 8k). Although the local rainfall is usually abundant, appropriate irrigation can help increase yields (Fig. 8l). According to the statistical yearbooks of Huzhou City and Hangzhou City in the past three years, the average yields were 4,565 kg/ha and 4,074 kg/ha, respectively, as shown by the red dashed lines in Fig. 8j-l.

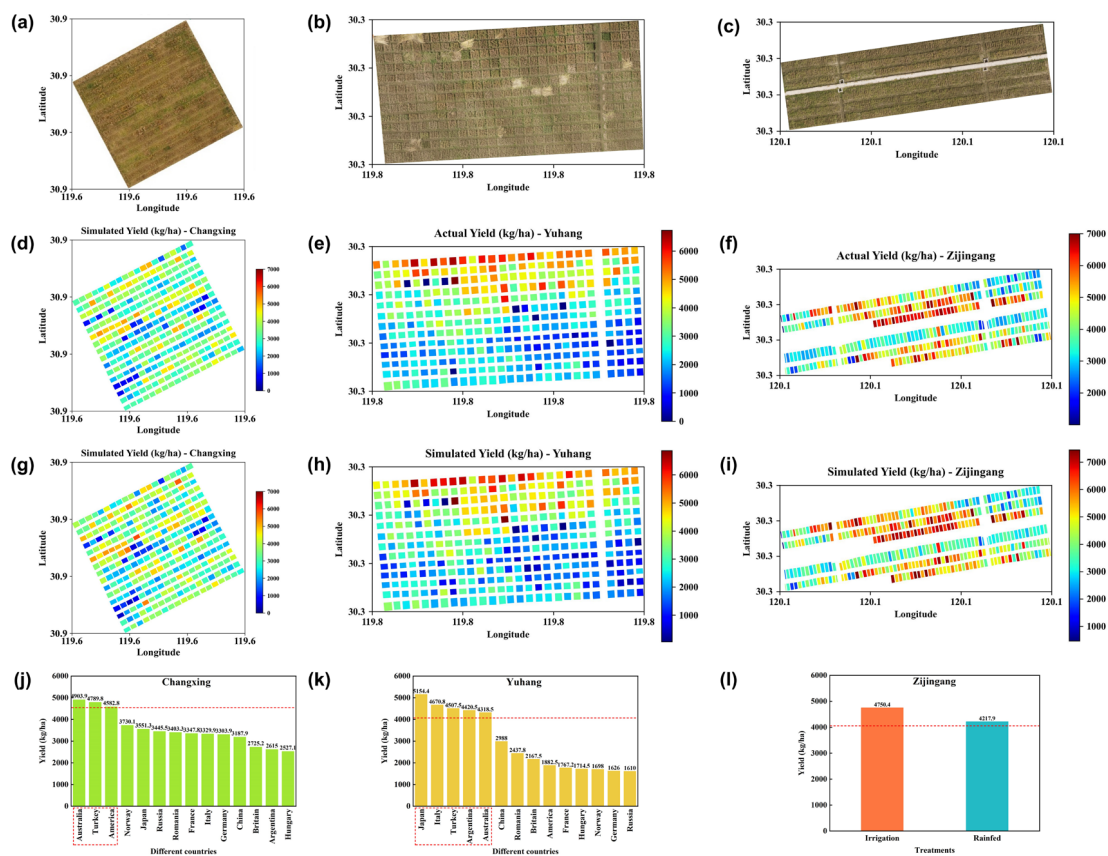

**Fig. 8.** Comparison of yield among different wheat breeding materials. (a-c) Experiment fields at maturity in Changxing, Yuhang, and Zijingang. (d) Predicted yields for the Changxing experiment using the WW-4VES data assimilation algorithm. (e,f) Measured yields in Yuhang and Zijingang. (g-i) Predicted yields using the hybrid method in Changxing, Yuhang, and Zijingang. (j,k) Average yield (kg/ha) of different wheat germplasms in different countries for the Changxing, Yuhang experiments. (l) Measured yield difference between irrigation and rain-fed for the Zijingang experiment. The red dashed line represents the average wheat yield (kg/ha) in the city over the past three years.

*4.3. ChatYield interactive yield predict*

The user first inputs a question (query) into ChatYield, which is then processed by LlamaIndex and the hybrid method respectively. The LLM comprehensively processes the new prompt from LlamaIndex and the output of the hybrid method. Finally, the Q/A system displays the final answer (response). Before using ChatYield, the user needs to calibrate and verify the WOFOST crop model locally to ensure the correct operation of subsequent data assimilation and TFT method.

We show examples of different user questions using ChatYield (red box in Fig. 9), GPT-4o (yellow box in Fig. 9), and pre-trained Llama3.2-3.2-3B-Instruct (blue box in Fig. 9). More examples can be found in Table S8. Using remote sensing time series data and yield database, we can quickly predict wheat yield and obtain the yield performance of this variety of wheat in other regions (Fig. 9a). If wheat breeders combine this with years of field experiments, they can more accurately find high-quality wheat varieties and analyze the stability of yields of different varieties. Compared with GPT-4o and pre-trained Llama3.2-3.2-3B-Instruct, it is obvious that we have enhanced the accuracy, flexibility, and timeliness of the generated results through RAG technology (Fig. 9b-d), because RAG technology provides a dynamic supplement mechanism for LLM. Furthermore, GPT-4o can answer partially correctly, while the pre-trained Llama3.2-3.2-3B-Instruct answers worse. ChatYield not only relies on the pre-trained knowledge of LLM, but also dynamically retrieves relevant information from the external yield database during the answer generation process, thereby incorporating the latest or more precise knowledge when generating answers.

**(a) Q:** Wheat variety name: Jimai 44. Date: 2022-12-27, DVD: 48, CIre: 0.324, CIg: 1.9, EVI: 0.247, GNDVI: 0.487, GRDVI: 0.28, kNDVI: 0.186, PSRI: 0.221, MSAVI: 0.237, MTCARI: 0.076, MTVI: 0.195, NLI: -0.232, NDRE: 0.139, NDVI: 0.434, NIRv: 0.107, NNIR: 0.575, OSAVI: 0.343, RVI: 0.395, SAVI: 0.265, SR: 2.532, SRre: 1.324, SIPI: 0.592, TVI: 8.448, VDVI: 0.096, VARI: -0.089, WDRVI: -0.596, CV: 0.09, CH: 0.2, LAI: 0.25, Date: 2023-02-16, DVD: 99, CIre: 0.72, CIg: 4.758, EVI: 0.701, GNDVI: 0.704, GRDVI: 0.542, kNDVI: 0.555, PSRI: 0.046, MSAVI: 0.661, MTCARI: -0.096, MTVI: 0.686, NLI: 0.624, NDRE: 0.265, NDVI: 0.791, NIRv: 0.399, NNIR: 0.775, OSAVI: 0.714, RVI: 0.117, SAVI: 0.628, SR: 8.55, SRre: 1.72, SIPI: 0.832, TVI: 27.882, VDVI: 0.298, VARI: 0.258, WDRVI: -0.078, CV: 0.14, CH: 0.21, LAI: 0.14, …, Date: 2023-05-13, DVD: 185, CIre: 0.362, CIg: 2.093, EVI: 0.278, GNDVI: 0.511, GRDVI: 0.326, kNDVI: 0.189, PSRI: 0.242, MSAVI: 0.28, MTCARI: 0.093, MTVI: 0.236, NLI: -0.121, NDRE: 0.153, NDVI: 0.437, NIRv: 0.134, NNIR: 0.583, OSAVI: 0.369, RVI: 0.391, SAVI: 0.302, SR: 2.555, SRre: 1.362, SIPI: 0.611, TVI: 10.376, VDVI: 0.089, VARI: -0.12, WDRVI: -0.593, CV: 0.84, CH: 0.84, LAI: 2.14, Date: 2023-05-19, DVD: 191, CIre: 0.297, CIg: 1.779, EVI: 0.187, GNDVI: 0.471, GRDVI: 0.247, kNDVI: 0.138, PSRI: 0.266, MSAVI: 0.179, MTCARI: 0.062, MTVI: 0.128, NLI: -0.386, NDRE: 0.129, NDVI: 0.373, NIRv: 0.075, NNIR: 0.55, OSAVI: 0.28, RVI: 0.457, SAVI: 0.207, SR: 2.19, SRre: 1.297, SIPI: 0.556, TVI: 5.81, VDVI: 0.053, VARI: -0.155, WDRVI: -0.641, CV: 0.52, CH: 0.82, LAI: 1.79. Please estimate the final wheat yield (kg/ha). Also tell me how this variety performs in yield elsewhere.

**A:** The estimated yield of Jimai 44 is 3441.44 kg/ha. In 2017-2018, it participated in the trial of the northern area of the Huanghuai winter wheat region of the National Wheat Variety Joint Research Program, with an average yield of 471.6 kg per mu, a 2.5% increase over the control Jimai 22; in the 2018-2019 trial, the average yield was 586.3 kg per mu, a 3.7% increase over the control Jimai 22; in the 2018-2019 production trial, the average yield was 603.2 kg per mu, a 2.8% increase over the control Jimai 22.

**(b) Q:** Please answer what are the wheat variety features of Jimai 44?

**A:** Semi-winter, full growth period 231.4 days, 1.3 days earlier than the control Jimai 22. The seedlings are semi-creeping, with wide and long leaves, light green leaves, and strong tillering ability. The plant height is 76.9 cm, the plant type is compact, and the resistance to lodging is good. The uniformity is good, the ear layer is neat, and the maturity is good. The ear is spindle-shaped, with long awns, white grains, horny grains, and good fullness. The number of ears per mu is 453,000, the number of grains per ear is 32.2, and the thousand-grain weight is 42.8 grams. Disease resistance identification: moderately resistant to powdery mildew, moderately susceptible to stripe rust and sheath blight, highly susceptible to leaf rust and fusarium head blight. Quality inspection: protein content 17.3%, 16.2%, wet gluten content 30.9%, 33.1%, stability time 26.9 minutes, 25.2 minutes, water absorption 60.1%, 65.4%, maximum tensile resistance 915.5 Rm.E.U., 574.8 Rm.E.U., tensile area 175 square centimeters, 109.8 square centimeters. The quality index reaches the standard of strong gluten wheat.

**Q:** Please answer what are the wheat variety features of Jimai 44?

**A:** Jimai 44 is a high-yielding wheat variety developed in China with several notable features, particularly for its agronomic and resistance traits. Below are the key characteristics: High Yield Potential: Jimai 44 is known for its excellent yield performance, especially under optimal growing conditions. Good Disease Resistance: It shows strong resistance to common wheat diseases such as powdery mildew, stripe rust, and Fusarium head blight (scab). This makes it suitable for regions prone to these diseases. Drought Resistance: Jimai 44 has some degree of drought tolerance, making it adaptable to environments with limited water availability. Strong Stem and Lodging Resistance: The variety has robust stems, which provide good lodging resistance, meaning the plants are less likely to fall over in adverse weather conditions. Grain Quality: Jimai 44 produces grains with good protein content, making it suitable for flour production, and it's also known for its good sedimentation value, which is important for bread-making quality. Early Maturity: This variety has a relatively short growth cycle, allowing it to be harvested earlier, which can be advantageous in double-cropping systems. Overall, Jimai 44 combines yield, disease resistance, and grain quality, making it a popular choice in several wheat-growing regions.

**Q:** Please answer what are the wheat variety features of Jimai 44?

**A:** After further research, I found that Jimai 44 is a wheat variety of China. According to my sources, here are some of its features: High yield potential, High resistance to Fusarium head blight, Good resistance to leaf rust, Good lodging resistance, Medium maturity days, Good baking quality, High protein content, Good gluten content. Please note that wheat varieties' traits and characteristics can vary depending on the source and location, so these features might not be exhaustive or universally applicable. If you have any further questions, I'll do my best to provide more information!

**(c) Q:** Please answer who is the applicant of the wheat variety Jimai 44? Who is its breeder? What is its variety source?

**A:** The applicant of Jimai 44 is the Crop Research Institute of Shandong Academy of Agricultural Sciences. The breeders are the Crop Research Institute of Shandong Academy of Agricultural Sciences and Shandong Luyan Agricultural Seed Co., Ltd. The variety source is the 954072/Jinan 17.

**Q:** Please answer who is the applicant of the wheat variety Jimai 44? Who is its breeder? What is its variety source?

**A:** Jimai 44 is a wheat variety developed in China. Here is the relevant information: Applicant: The applicant for the variety Jimai 44 is the Shandong Academy of Agricultural Sciences (SAAS), specifically through its Crop Research Institute. Breeder: The variety was bred by Professor Ma Wuying and her team at the Shandong Academy of Agricultural Sciences. Variety Source: Jimai 44 is a result of crossbreeding between the varieties Jimai 38 and Shan 229, combining traits to enhance yield, disease resistance, and adaptability to local environmental conditions. This variety is popular for its high yield potential, good disease resistance, and adaptability to various growing conditions.

**Q:** Please answer who is the applicant of the wheat variety Jimai 44? Who is its breeder? What is its variety source?

**A:** I don't have the specific information about the wheat variety Jimai 44, including who its applicant and breeders are or what it originated from.

**(d) Q:** Please answer the key points of cultivation techniques for the wheat variety Jimai 44 and what are the suitable planting areas?

**A:** Suitable for planting in high-fertilizer plots of more than 500 kg, suitable sowing period in early and mid-October, basic seedlings per mu are about 150,000. Carry out winter irrigation and spring hoeing to cultivate strong seedlings. Apply 10 kg of urea per mu during the jointing period, timely prevent and control aphids, fusarium head blight, rust and weeds, and harvest in time. It is suitable for planting in the entire Shandong Province in the northern water areas of the Huanghuai winter wheat region, the southern part of Baoding City and Cangzhou City in Hebei Province and the areas to the south, and the basin irrigation areas of Yuncheng and Linfen City in Shanxi Province.

**Q:** Please answer the key points of cultivation techniques for the wheat variety Jimai 44 and what are the suitable planting areas?

**A:** Key cultivation techniques and suitable planting areas for Jimai 44: Key Cultivation Techniques: Sowing Time: Mid to late October in northern winter wheat regions. Sowing Density: 220,000-250,000 plants/ha in well-watered areas, 180,000-200,000 plants/ha in dry areas. Fertilization: Apply balanced nitrogen, phosphorus, and potassium, with additional nitrogen during jointing. Irrigation: Focus on jointing, booting, and grain filling stages. Pest Control: Regular monitoring for diseases like rust and powdery mildew. Harvesting: Harvest at 14-15% grain moisture content. Suitable Planting Areas: Best suited for Yellow and Huai River Valley and Northern China, especially Shandong, Hebei, Henan, and parts of Anhui and Shaanxi.

**Q:** Please answer the key points of cultivation techniques for the wheat variety Jimai 44 and what are the suitable planting areas?

**A:** I don't have specific technical information on the cultivation techniques for Jimai 44 wheat variety. However, I can provide general information on wheat cultivation and the factors that affect the suitability of different wheat varieties. Jimai 44 is widely cultivated in: 1. China: Mainly in the provinces of Xinjiang, Ningxia, Gansu, Qinghai, and Inner Mongolia. 2. Developing regions: Countries with similar climate and soil conditions, such as countries in Central Asia, Eastern Europe, and parts of Africa, can also consider cultivating Jimai 44.

**Fig. 9.** Examples of using ChatYield (red box), GPT-4o (yellow box), and Llama3.2-3.2-3B-Instruct (blue box). (a) Yield prediction query for a specific wheat variety and the yield performance of the variety in other regions. (b) Characteristics of a specific wheat variety, including days in the full growth period, plant height, grain color, disease resistance, protein content, etc. (c) Applicants and breeders of a specific wheat variety and which varieties have bred that variety. (d) Key points of cultivation techniques for a specific wheat variety and suitable planting areas.

## 5. Discussion

*5.1. Advantages of hybrid method and tool*

Compared with process-based model-driven crop growth models, data-driven deep learning models and other hybrid methods, the hybrid method of this study has advantages in prediction accuracy. First, the hybrid method significantly outperforms the traditional calibrated WOFOST model in terms of crop yield prediction performance. This improvement is mainly attributed to the incorporation of deep learning, which can exploit the complex interrelationships between crop yield and other phenotyping (e.g., CV, CH). However, process-based models may not be able to fully describe these relationships [73]. The accuracy of simulations can be further improved by leveraging the knowledge constraints of crop models and the flexibility of deep learning [74]. At the same time, deep learning can integrate multiple data sources and effectively incorporate crop model predictions from all time alongside remote sensing data [9]. Then, the hybrid method shows superior performance than standalone deep learning models (GRU, LSTM, and Transformer) on different experiments. Unlike large-scale (global, national) deep learning models that rarely consider changes in agricultural practices [75], [76], [77], the hybrid method utilize process-based crop model that can be adapted to actual field management. This adaptability improves the accuracy of yield predictions, making this method valuable for decision-making in agricultural management practices. In addition, we incorporate remote sensing data assimilation into the hybrid method of this study to improve the performance of the crop model by providing more accurate and timely input data [19]. Furthermore, the integration of remote sensing data provides information on crop growth processes, thereby predicting yields more reliably than previous studies that used only the output variables of crop models as predictors.

For the yield prediction tool ChatYield, users can complete operations through natural language conversations without having to master complex menus or instructions, which reduces learning costs and improves user experience. ChatYield can simultaneously generate text and retrieve the latest information from external databases and the networks in real time, ensuring that the answers provided are accurate and timely. However, the data update cycle of static storage and display webs or desktop applications is relatively long and the information may be delayed [27], [78], [79]. ChatYield can handle information needs in multiple disciplines and fields, not limited to yield prediction and phenotyping retrieval, breaking the limitations of traditional webs or desktop applications focusing on a single function [80] and reducing the frequency of users switching between different applications. ChatYield can remember the conversation context of the user, provide personalized suggestions and continuous conversation experience, while traditional webs and desktop applications mostly require users to repeatedly enter the same type of data and cannot continuously track user needs [81]. ChatYield combines generation and retrieval functions to quickly provide solutions to complex problems, reduce the workload of users to manually search and filter information, and improve the efficiency of task completion. Traditional breeding tools often require users to search and analyze on their own [26]. ChatYield is highly extensible, capable of integrating external data sources or systems through plugins and APIs, allowing it to adapt to evolving needs. However, breeding software tends to have limited extensibility, with relatively long development cycles and higher associated costs. [82].

*5.2. Key factors contributing to the hybrid method*

This study used SHAP (Shapley Additive exPlanations) analysis [83] to determine feature importance and gain insight into the relationship between feature values and yield prediction impact. We extracted feature importance from the SHAP method, which is based on the magnitude of feature attributions and is computed as the average absolute Shapley value. It differs from permutation feature importance which is based on the decrease in model performance.

From the SHAP analysis, we determined feature importance and feature directionality. The latter can be visualized from Fig. 10a-c by comparing feature values with SHAP values: whether a positive or negative effect on prediction is related to feature values. Overall, the top three most important features in each of the three experiments were physiological parameters (TWRT, TAGP) from the crop model and VIs (MTCARI, NLI, NIRv, SIPI) from remote sensing with large feature importance scores. However, the importance scores of morphological characteristics (LAI, CV, and CH) were lower compared with physiological parameters and VI. Likewise, it is evident that the influence of many features on model predictions is mixed (with unclear feature directionality). Moreover, the importance of different features in different growth stages varies significantly and changes over time, which is similar to other study [21]. Unlike the features in the late growth stages that usually show the highest importance to yield [84], we found that some features may be highly important at all stages of the entire growth period (Fig. 10d-f). This mixed influence of features on prediction indicates that the relationship between features and yield is nonlinear [13]. The importance of some VIs (e.g., EVI, kNDVI) is relatively low, and they can be excluded in subsequent applications to reduce the amount of calculation and have almost no effect on the yield prediction results [85]. Since the wheat in three experiments of our study was grown in a natural environment with weed competition, feature importance scores may be influenced by specific growth conditions, crop varieties, and agricultural management practices. As a result, the ranking of feature importance may not be entirely consistent, and the conclusion similar to other studies [86], [87], [88].

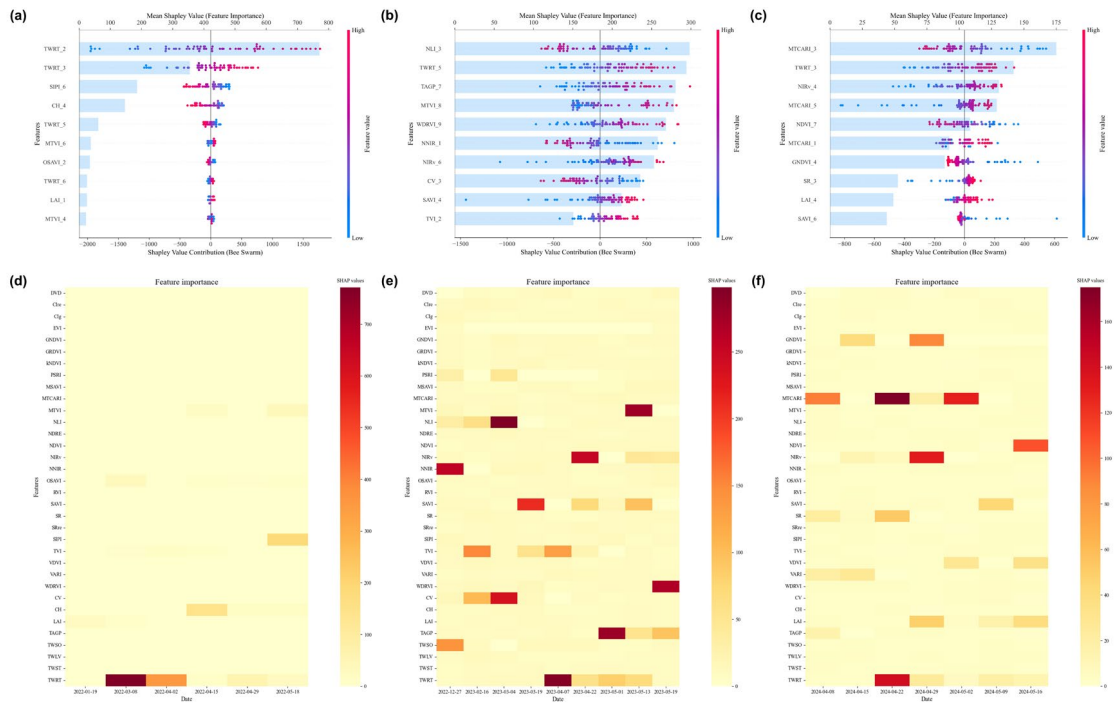

**Fig. 10.** The hybrid method was run on the test set for three experiments respectively. (a-c) are dual-axis charts, comprising SHAP value bee-swarm plots and feature importance bar charts. These charts display contribution of each feature to model predictions and its average importance. In the bee-swarm plots, the X-axis below shows the distribution of SHAP values for each feature, with color indicating feature value magnitude. In the bar charts, the X-axis above displays the average SHAP value for each feature, serving as an importance ranking. Each point corresponds to a model prediction. The number after the feature name indicates its order in the sequence of collected data. (d-f) Feature importance heat map based on SHAP value. The color depth represents the size of the SHAP value. The darker the color, the more important the feature is at that time point, which means the greater its contribution to yield prediction.

*5.3. Limitations and future directions*

This study demonstrates that the hybrid method is promising in improving crop yield prediction performance, outperforming traditional crop models and classic deep learning models, and the yield prediction interactive tool ChatYield is convenient and user-friendly, capable of integrating multi-source data and supporting breeding decisions. Nevertheless, the limitations of this study must be acknowledged and suggestions for future research are provided. The performance of the hybrid method depends largely on data quality and availability, such as meteorological data, agricultural management practices data, and remote sensing data. Similar to other studies, there are differences in planting environments and wheat varieties and limited data used in the studies, which may affect the prediction performance of the hybrid method [18], [21]. In addition, using VI to invert LAI is a common method, which may encounter the saturation problem of VI, spectral confusion, background influence [89], and the calibration effect of model parameters. However, this method is simple and convenient, and can obtain more accurate LAI inversion results after parameter calibration [90], [91]. For the CH and CV inversion of this study, the biggest possible influence is weeds. Direct competition for resources is generally considered to be the main mechanism of yield loss caused by weeds [92]. Nevertheless, this study focuses on wheat high-yield breeding under natural weed competition, and the results demonstrate a satisfactory yield prediction performance.

Future research should focus on improving data quality and scale and exploring alternative data sources to enhance the robustness of the model. Although this study conducted wheat yield predictions in three regions, it still needs to be calibrated and verified based on local environmental and meteorological data when it is actually applied in other regions. Nevertheless, considering region-specific factors such as agronomic practices, climatic conditions, and crop management strategies, the hybrid approach may be a more effective method for yield prediction. Crop breeding involves cross-domain, multimodal, and large-scale data [93]. The yield prediction tool we developed alone is not enough to support complete breeding services, and new functions need to be supplemented and improved in the future to better support breeding decisions.

## 6. Conclusion

This study integrates a newly developed hybrid method (remote sensing data assimilation and deep learning) with a large language model to achieve interactive and accurate predictions of wheat breeding yields. Firstly, the newly designed WW-4VES data assimilation algorithm is used to assimilate LAI data into the WOFOST model. Then, selected outputs from the assimilation process, along with remote sensing inversion results, are used to drive the time-series TFT model for wheat yield prediction. Finally, based on the hybrid method and leveraging LLM with RAG technology, we developed an interactive yield prediction Web tool that is user-friendly and supports sustainable data updates. This tool integrates multi-source data to support breeding decision-making. This study aims to accelerate the identification of high-yield materials in the breeding process, enhance breeding efficiency, and enable more scientific and smart breeding decision-making.

**CRediT authorship contribution statement**

**Guofeng Yang:** Conceptualization, Data curation, Formal analysis, Investigation, Methodology, Validation, Visualization, Writing – original draft. **Nanfei Jin:** Investigation. **Wenjie Ai:** Investigation. **Zhonghua Zheng:** Writing – review & editing. **Yuhong He:** Writing – review & editing. **Yong He:** Project administration, Resources.

**Declaration of Competing Interest**

The authors declare that they have no known competing financial interests or personal relationships that could have appeared to influence the work reported in this paper.

**Acknowledgments**

Thanks to Professor Xianchun Xia and Dr. Jindong Liu (Chinese Academy of Agricultural Sciences) for providing wheat germplasm resources. Thanks to the Zhejiang University wheat breeding group for providing the advanced lines for wheat breeding. Thanks to the members of the Digital Phenotyping Research Group of Zhejiang University for their help in data collection. This research was funded by the National Key Research and Development Program of China (2021YFD2000102) and Zhejiang Provincial Key R&D Program of China (Grant No. 2022C02013).

**Appendix A. Supplementary data**

Supplementary material